\pdfoutput=1 

\documentclass[10pt,twocolumn,letterpaper]{article}

\usepackage[pagenumbers]{cvpr} 
\usepackage[accsupp]{axessibility}  

\usepackage{graphicx}
\usepackage{amsmath}
\usepackage{amssymb}
\usepackage{booktabs}
\usepackage{amsmath}
\usepackage{multirow}
\usepackage{colortbl}

\usepackage{enumitem}
\setitemize[1]{itemsep=0pt,partopsep=0pt,parsep=\parskip,topsep=0pt}

\usepackage[pagebackref,breaklinks,colorlinks]{hyperref}

\usepackage[capitalize]{cleveref}
\crefname{section}{Sec.}{Secs.}
\Crefname{section}{Section}{Sections}
\Crefname{table}{Table}{Tables}
\crefname{table}{Tab.}{Tabs.}

\def\cvprPaperID{9466} 
\def\confName{CVPR}
\def\confYear{2023}

\makeatletter
\def\thanks#1{\protected@xdef\@thanks{\@thanks
        \protect\footnotetext{#1}}}
\makeatother
\begin{document}

\title{Video Event Restoration Based on Keyframes for Video Anomaly Detection}

\author{Zhiwei Yang$^1$, Jing Liu$^1$$^\dagger$\thanks{$^\dagger$Corresponding authors.}, Zhaoyang Wu$^1$, Peng Wu$^2$$^\dagger$, Xiaotao Liu$^1$\\ 
$^1$Guangzhou Institute of Technology, Xidian University, Guangzhou, China\\ 
$^2$School of Computer Science, Northwestern Polytechnical University, Xi'an, China\\
{\tt\small \{zwyang97, neouma, 15191737495\}@163.com, xdwupeng@gmail.com, xtliu@xidian.edu.cn}
}
\maketitle

\begin{abstract}
   Video anomaly detection (VAD) is a significant computer vision problem. Existing deep neural network (DNN) based VAD methods mostly follow the route of frame reconstruction or frame prediction. However, the lack of mining and learning of higher-level visual features and temporal context relationships in videos limits the further performance of these two approaches. Inspired by video codec theory, we introduce a brand-new VAD paradigm to break through these limitations: First, we propose a new task of video event restoration based on keyframes. Encouraging DNN to infer missing multiple frames based on video keyframes so as to restore a video event, which can more effectively motivate DNN to mine and learn potential higher-level visual features and comprehensive temporal context relationships in the video. To this end, we propose a novel \textbf{U}-shaped \textbf{S}win \textbf{T}ransformer \textbf{N}etwork with \textbf{D}ual \textbf{S}kip \textbf{C}onnections (\textbf{USTN-DSC}) for video event restoration, where a cross-attention and a temporal upsampling residual skip connection are introduced to further assist in restoring complex static and dynamic motion object features in the video. In addition, we propose a simple and effective adjacent frame difference loss to constrain the motion consistency of the video sequence. Extensive experiments on benchmarks demonstrate that USTN-DSC outperforms most existing methods, validating the effectiveness of our method.
\end{abstract}

\section{Introduction}
\label{sec:intro}

Video anomaly detection (VAD) is a hot but challenging research topic in the field of computer vision. One of the most challenging aspects comes from the scarcity of anomalous samples, which hinders us from learning anomalous behavior patterns from the samples. As a result, it is hard for supervised methods to show their abilities, as unsupervised methods are by far the most widely explored direction \cite{adam2008robust-1,benezeth2009abnormal-2, sabokrou2015real-4, zhai2016deep-5,gong2019memorizing-6,liu2018future-9,hasan2016learning-8,ye2019anopcn-27,luo2017revisit-40,park2020learning-45,zhong2022cascade-46,lv2021learning-47,wang2021robust-48,cai2021appearance-49,yang2022dynamic-28,luo2017remembering-25,tudor2017unmasking-52,nguyen2019anomaly-53,zhou2019anomalynet-54,wu2019deep-55}. To determine whether an abnormal event occurs under unsupervised, a general approach is to model a regular pattern based on normal samples, leaving samples that are inconsistent with this pattern as anomalies. 
\begin{figure}[t]
  \centering
   \includegraphics[width=0.85\linewidth]{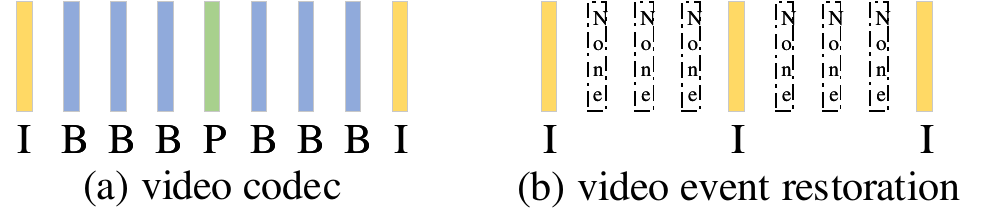}
   \caption{Video codec $vs.$ video event restoration.
  Compared to video decoding based on explicit motion information, our proposed video event restoration task encourages DNN to automatically mine and learn the implied spatio-temporal relationships from several frames with key appearance and temporal transition information to restore the entire video event.}
   \label{fig:video_codec}
\end{figure}

Under the unsupervised framework, reconstruction and prediction-based approaches are two of the most representative paradigms \cite{hasan2016learning-8,liu2018future-9} for VAD in the current era of deep neural networks (DNN). The reconstruction-based approaches typically use deep autoencoder (DAE) to learn to reconstruct an input frame and regard a frame with large reconstruction errors as anomalous. However, the powerful generalization capabilities of DAE enable well reconstruction of anomalous as well \cite{gong2019memorizing-6}. This is attributed to the simple task of reconstructing input frames, where DAE only memorizes low-level details rather than understanding high-level semantics \cite{larsen2016autoencoding-7}. The idea of the prediction-based approach assumes that anomalous events are unpredictable and then builds a model that can predict future frames according to past frames, and poor predictions of future frames indicate the presence of anomalies \cite{liu2018future-9}. The prediction-based approaches further consider short-term temporal relationships to model appearance and motion patterns. However, due to the appearance and motion variations of adjacent frames being minimal, the predictor can predict the future frame better based on the past few frames, even for anomalous samples.
The inherent lack of capability of these two modeling paradigms in mining higher-level visual features and comprehensive temporal context relationships limits their further performance improvement. 

In this paper, we aim to explore better methods for mining complex spatio-temporal relationships in the video. Inspired by video codec theory \cite{ladune2022aivc-10,liu2010key-61}, we propose a brand-new VAD paradigm: video event restoration based on keyframes for VAD. In the video codec \cite{liu2010key-61}, three types of frames are utilized, namely I-frame, P-frame, and B-frame. I-frame contains the complete appearance information of the current frame, which is called a keyframe. P-frame contains the motion difference information from the previous frame, and B-frame contains the motion difference between the previous and next frames. Based on these three types of frames containing explicit appearance and motion relative relationships, the video can be decoded. Inspired by this process, our idea sprang up: It should be also theoretically feasible if we give keyframes that contain only implicit relative relations between appearance and motion, and then encourage DNN to explore and mine the potential spatio-temporal variation relationships therein to infer the missing multiple frames for video event restoration. To motivate DNN to explore and learn spatio-temporal evolutionary relationships in the video actively, we do not provide frames like P-frames or B-frames that contain explicit motion information as input cues. This task is extremely challenging compared to reconstruction and prediction-based tasks, because DNN must learn to infer the missing multiple frames based on keyframes only. This requires a strong regularity and temporal correlation of the event in the video for a better restoration. On the contrary, video restoration will be poor for anomalous events that are irregular and random. Under this assumption, our proposed keyframes-based video event restoration task can be exactly applicable to VAD. \cref{fig:video_codec} compares the video codec and video event restoration task with an illustration.

To perform this challenging task, we propose a novel \textbf{U}-shaped \textbf{S}win \textbf{T}ransformer \textbf{N}etwork with \textbf{D}ual \textbf{S}kip \textbf{C}onnections (\textbf{USTN-DSC}) for video event restoration based on keyframes. USTN-DSC follows the classic U-Net \cite{ronneberger2015u-12} architecture design, where its backbone consists of multiple layers of swin transformer (ST) \cite{liu2021swin-11} blocks. 
Furthermore, to cope with the complex motion patterns in the video so as to better restore the video event, we build dual skip connections in USTN-DSC. Specifically, we introduce a cross-attention and a temporal upsampling residual skip connection to further assist in restoring complex dynamic and static motion object features in the video. In addition, to ensure that the restored video sequence has the consistency of temporal variation with the real video sequence, we propose a simple and effective adjacent frame difference (AFD) loss. Compared with the commonly used optical flow constraint loss \cite{liu2018future-9}, AFD loss is simpler to be calculated while having comparable constraint effectiveness.

The main contributions are summarized as follows:
\begin{itemize}
\item We introduce a brand-new video anomaly detection paradigm that is to restore the video event based on keyframes, which can more effectively mine and learn higher-level visual features and comprehensive temporal context relationships in the video.
\item We introduce a novel model called USTN-DSC for video event restoration, where a cross-attention and a temporal upsampling residual skip connection are introduced to further assist in restoring complex dynamic and static motion object features in the video.
\item We propose a simple and effective AFD loss to constrain the motion consistency of the video sequence.
\item USTN-DSC outperforms most existing methods on three benchmarks, and extensive ablation experiments demonstrate the effectiveness of USTN-DSC.
\end{itemize}

\section{Related Work}
\label{sec:formatting}

\subsection{Video Anomaly Detection}

Over the past years, extensive works have been devoted to solving the VAD problem \cite{gong2019memorizing-6,liu2018future-9,hasan2016learning-8,ye2019anopcn-27,luo2017revisit-40,park2020learning-45,zhong2022cascade-46,lv2021learning-47,wang2021robust-48,cai2021appearance-49,yang2022dynamic-28,luo2017remembering-25,tudor2017unmasking-52,nguyen2019anomaly-53,zhou2019anomalynet-54,wu2019deep-55,wu2020not-58,yang2021bidirectional-59}, which can be mainly categorized into two main groups based on traditional methods and deep neural network-based methods.

\noindent\textbf{VAD based on traditional methods.} Traditional VAD methods mainly utilize statistical models based on hand-extracted features or classical machine learning techniques. For example, Adam \etal in \cite{adam2008robust-1} characterized the normal local histograms of optical based on statistical monitoring of low-level observations at multiple spatial locations. Kim and Grauman \cite{kim2009observe-14} modeled the local optical flow pattern with a mixture of probabilistic principal component analyzers and trained a space-time markov random field to infer abnormalities. Cong \etal in \cite{cong2011sparse-15} introduced a sparse reconstruction cost over the normal dictionary to measure the normality of testing samples. Although these traditional methods achieve better results in specific scenarios, their performance in some complex scenarios is severely constrained owing to poor feature representation capabilities. 

\noindent\textbf{VAD based on deep learning methods.} With deep learning techniques flourishing in various fields \cite{krizhevsky2017imagenet-17,he2016deep-18,redmon2016you-19,ren2015faster-20,nie2020deep-22,shao2022exploiting-60}, anomaly detection methods based on deep learning have also been widely studied. The most prevalent of these methods are frame reconstruction and frame prediction. For example, Hasan \etal in \cite{hasan2016learning-8} used the extracted features as input to a fully connected neural network-based autoencoder to learn the temporal regularity in the video. A regularity score was calculated according to the reconstruction error and used to determine whether an abnormality occurs. Xu \etal in \cite{xu2017detecting-24} proposed a stacked denoising autoencoder to separately learn both the appearance and the motion features. Liu \etal in \cite{liu2018future-26} presented a video anomaly detection method that predicts the future frame with the U-Net architecture \cite{ronneberger2015u-12}. Yang \etal in \cite{yang2022dynamic-28} introduced a dynamic local aggregation network with adaptive clusterer for enhancing the representation capability of normal prototypes in the prediction paradigm. Although the approaches of frame reconstruction and frame prediction currently show promising results, the lack of mining and learning of higher-level visual features and comprehensive temporal context relationships hinder further performance improvement. 
\subsection{Video Restoration}
Video restoration, such as video super-resolution \cite{caballero2017real-29,haris2020space-30,geng2022rstt-31}, denoising \cite{hyun2017online-33}, deblurring \cite{sheth2021unsupervised-34}, and inpainting \cite{zou2021progressive-35}, has become a popular research topic in recent years. It aims to restore a clear and high quality video from a degraded low quality video. For example, Geng \etal in \cite{geng2022rstt-31} proposed a real-time spatial temporal transformer to effectively incorporate all the spatial and temporal information for synthesizing high frame rate and high resolution videos. Kim \etal in \cite{hyun2017online-33} proposed a fast online video deblurring method by efficiently increasing the receptive field of the network without adding a computational overhead to handle large motion blurs. Sheth \etal in \cite{sheth2021unsupervised-34} proposed an unsupervised deep video denoiser, a convolutional neural network architecture designed to be trained exclusively with noisy data. Zou \etal in \cite{zou2021progressive-35} proposed a progressive temporal feature alignment network for video inpainting, which fills the missing regions by making use of both temporal convolution and optical flow. 

Difference from these existing video restoration tasks, in this paper, we propose a new video restoration task that restores the video event based on keyframes. This task is more challenging compared to existing video restoration tasks because the missing multiple frames result in the discontinuity of temporal clues. This requires a strong regularity and temporal correlation of the events in the video for a better restoration. On the contrary, there will be a large restoration error for anomalous events, as it is irregular and random. Therefore, the task of restoring the video event based on keyframes can be well applied to VAD. 

\begin{figure*}[t]
  \centering
   \includegraphics[width=0.8\linewidth]{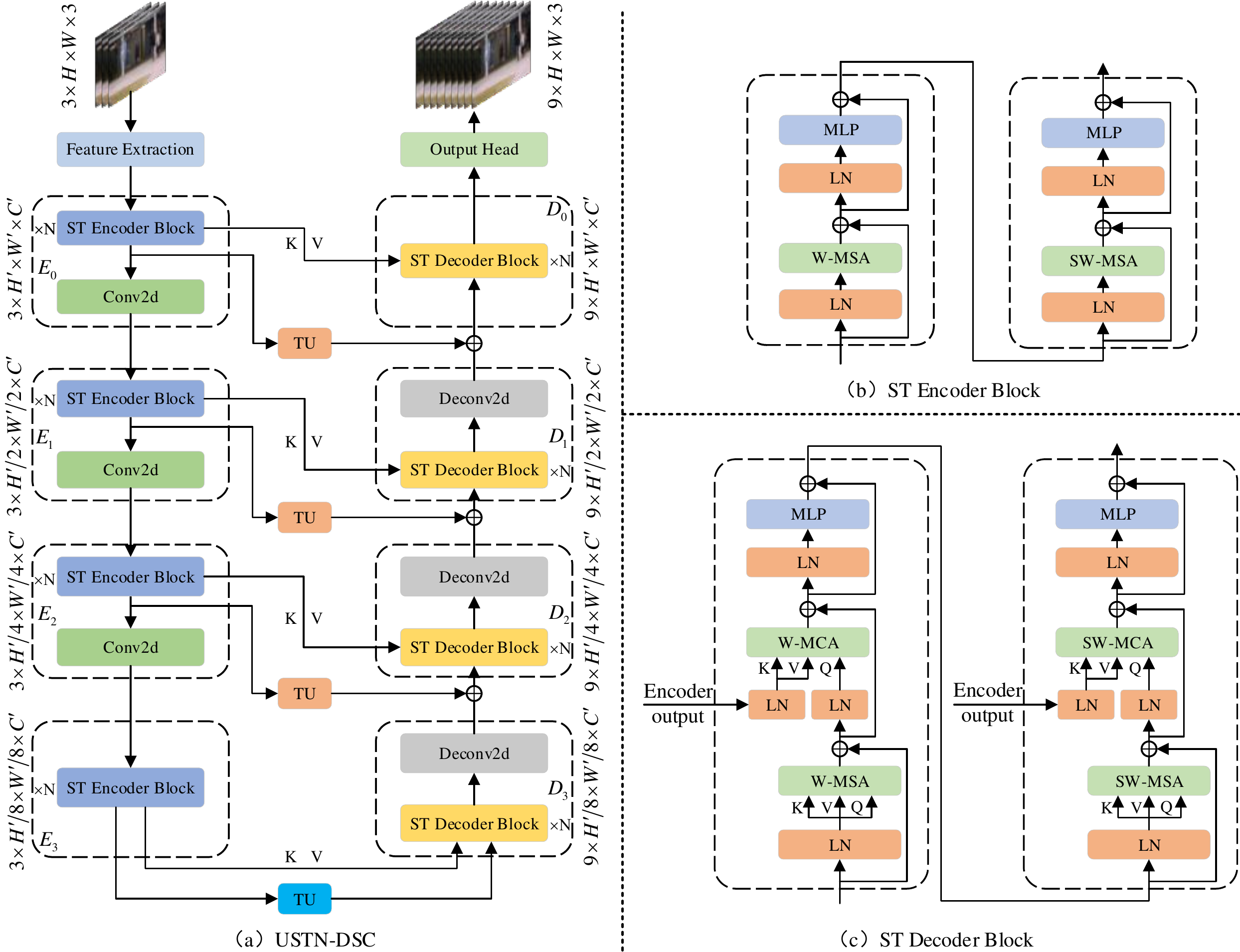}
   \caption{The overview architecture of USTN-DSC. (a) The complete network structure of USTN-DSC. TU represents the temporal upsampling module. (b) The structure of ST Encoder Block. LN represents Layer Normalization. W-MSA and SW-MSA are multi-head self-attention modules with regular and shifted windowing configurations, respectively. (c) The structure of ST Decoder Block. W-MCA and SW-MCA are multi-head cross-attention modules with regular and shifted windowing configurations, respectively.}
   \label{fig:main}
\end{figure*}

\section{Method}

In the unsupervised VAD framework, most approaches devote to designing models that characterize normal behavior patterns and consider deviations from them as anomaly classes.
To explore a superior approach to modeling normal behavior, inspired by video codec theory \cite{ladune2022aivc-10}, we propose a novel normal behavior modeling paradigm for VAD: Restore video event based on Keyframes to detect anomalies. Concretely, given a video sequence of length $T$, we take $L$ keyframes in the video sequence as input, aiming to recover the missing $T-L$ frames according to these keyframes. Compared with reconstruction or prediction tasks, it is more challenging but better to motivate DNN to mine and learn the higher-level visual features and comprehensive temporal context relationships in video sequences. To meet this challenging task, we propose a novel model called USTN-DSC for video event restoration. Next, we will describe the architecture and workflow of USTN-DSC in detail. 

\subsection{Network Overview}

USTN-DSC follows the U-Net \cite{ronneberger2015u-12} architecture design, which consists of four parts: a feature extractor, an encoder, a decoder, and an output head. The feature extractor and output head mainly consist of 2D convolutional layers, and the encoder and decoder are a combination of swin transformer (ST) \cite{liu2021swin-11} and 2D convolutional layers.

\cref{fig:main} illustrates the overall framework of USTN-DSC. Given a video sequence  $S=\left\{I_t|I_t\in\mathbb{R}^{H\times\ W\times\ C}\right\}_{t=1}^T $, where $T$ denotes the length of the video sequence and $H, W, C$ denote the height, width and number of channels of a video frame, respectively. Following the video codec theory \cite{liu2010key-61}, we first select the first and last frames of the video sequence as keyframes. 
To encourage USTN-DSC to automatically learn potential appearance and motion evolution relationships in the video sequence, instead of giving P-frames and B-frames with explicit motion information, we take the intermediate frame of video sequences as temporal transition frame for spatio-temporal relationship development. Therefore, we take $I_1$, $I_{(T-1)/2+1}$, and $I_T$ of the video sequence $S$ as the three keyframes of the input, and stack them up in chronological order as $X\in\mathbb{R}^{3 \times H \times W \times C}$. Then, the input $X$ is first processed by the feature extraction module $F_e$ for initial feature extraction and dimensionality reduction. Next is the encoding part, and the encoder consists of four stages, denoted by $E_n, n=0,1,2,3$ , each of which is a stack of ST encoder block $E_n^{ST}$ followed by a convolution layer $\varphi_n$ , except for the final stage $E_3$. Symmetrically with the encoder, the decoder is denoted by $D_n, n=0,1,2,3$, each of which is a stack of ST decoder block $D_n^{ST}$ followed by a deconvolution layer ${\varphi}_n^{-1}$, except for the first stage $D_0$. Finally, the output of the decoder is further transformed by the output head $F_{out}$ to obtain the restored video sequence $\hat{S}=\left\{{I}_t|{I}_t\in\mathbb{R}^{H\times\ W\times\ C}\right\}_{t=1}^T$.

Specifically, in USTN-DSC, $F_e$ first extracts the features $f_X\in\mathbb{R}^{3 \times H^{'} \times W^{'} \times C^{'}}$ from $X$. Then, $E_0^{ST}$ first takes $f_X$ as input and obtains $f^e_0=E_0^{ST}(f_X)\in\mathbb{R}^{3 \times H^{'} \times W^{'} \times C^{'}}$ and then follows a convolutional layer to obtain $e_0={\varphi_0}(f^e_0)\in\mathbb{R}^{3 \times H^{'}/2 \times W^{'}/2 \times C^{'}}$. Subsequently, we have
\begin{equation}
\begin{cases}
f^e_n=E_n^{ST}(e_{n-1}),  & n=1,2,3 \\
e_n=\varphi_n(f^e_n),     & n=1,2\\
e_3=f^e_3
\end{cases}.
\label{eq:eq1}
\end{equation}
During the encoding phase, each $e_n$ corresponds to the output features maps of the three keyframes, i.e. $e_n\equiv \left\{e^1_n,e_n^{(T-1)/2+1},e^T_{n} \right\}$. After obtaining the final output $e_3$ from the encoder, a temporal upsampling (TU) module located at the bottleneck generates the initial features $q_{0}=TU(e_3)$ based on $e_3$ for subsequent restoration of missing frames. Then, $q_0$ is further divided into two parts $q^f_{0}$ and $q^b_{0}$. $q^f_{0}$ represents the initial features of the missing frames between $I_1$ and $I_{(T-1)/2+1}$, and $q^b_{0}$ represents the one between $I_{(T-1)/2+1}$ and  $I_T$. For the features corresponding to the timestamps of $I_1$, $I_{(T-1)/2+1}$, and $I_T$, we directly keep following the ones on the corresponding time points in $e_3$. Eventually, the prototype features used for decoder input are represented as $Q:=(e^1_3,q^f_0,e^{(T-1)/2+1}_3,q^b_0,e^T_3)$. 

Next, we move on to the decoding phase. First, $D_3^{ST}$ in $D_3$ takes $e_3$ and $Q$ as input and obtains $f^d_3=D_3^{ST}(e_3,Q)$ and then follows a deconvolution layer to obtain $d_3={\varphi_3^{-1}}(f^d_3)$. It is noted that $D_n^{ST}$ here differs from the original ST block which only compute self-attention, we construct a skip connection from the encoder to compute cross-attention, which is the first channel of the dual skip connections. Next, we construct the second skip connection. The features $e_2$ from the encoder are temporally upsampled into $e^u_2=TU(e_2)$ using the TU module. Then, similar to the synthesis process of the prototype features $Q$, $e^u_2$ and $e_2$ are temporally dimensionally concatenated to obtain the combined features $e^r_2$. Then,  $e^r_2$ are added to $d_3$ in the form of residual and fed into $D_2^{ST}$ followed by a deconvolution layer ${\varphi_2}^{-1}$. The operation of the subsequent layers is similar and we formulate them as follows:
\begin{equation}
\begin{cases}
d_3={\varphi_3^{-1}}(D_3^{ST}(e_3,Q)), \\
e^r_n=TU(e_n)\bigcup e_n,     & n=0,1,2\\
d_n={\varphi_n^{-1}}(D_n^{ST}(e_n,d_{n+1}+e^r_n)) & n=1,2\\
d_0 = D_0^{ST}(e_0,d_1+e^r_0)
\end{cases}.
\label{eq:eq2}
\end{equation}
Finally, the output features $d_0$ are transformed into the restored video sequence $\hat{S}=\left\{{I}_t|{I}_t\in\mathbb{R}^{H\times\ W\times\ C}\right\}_{t=1}^T $ by the output head $F_{out}$. we show the specific structure of the $F_e$ and $F_{out}$ in the supplementary materials.


\subsection{Encoder}
The encoder of USTN-DSC mainly consists of stacked multiple ST encoder blocks followed by a convolutional layer. The detailed structure of the ST encoder block is shown in \cref{fig:main}(b). 
Although we deal with video data here, we do not use the division of the 3D shifted windows as in the video swin transformer \cite{liu2022video-36} for capturing the temporal relationships, as this is a bit trivial for the input of only three frames. In order to use a more simple way to equip the ST with the ability to learn long-range spatio-temporal dependencies, we calculate the local window attention by combining all the windows on the space corresponding to the current window simultaneously. Specifically, 
given an input video sequence of size $3\times H^{'}\times W^{'}\times C^{'}$, a ST block partitions it into non-overlapping windows of size $3\times \left \lceil\frac{H^{'}}{M}\right \rceil \times \left \lceil\frac{W^{'}}{M}\right \rceil \times C^{'}$. Here we choose, $H^{'}=W^{'}=64$, $M=4$ and $C^{'}=96$. Then, we reshape the video frame features with divided windows into $\frac{H^{'}W^{'}}{M^2}\times 3M^2 \times C^{'}$, where $\frac{H^{'}W^{'}}{M^2}$ is the total number of windows. Next, the reshaped features are first layer normalized (LN) and followed by a window-based multi-head self-attention (W-MSA) \cite{liu2021swin-11} to compute the local attention of each window. Immediately after, a multi-layer perception (MLP) follows another LN layer for further features transformations. Then, an additional ST block with shifted window-based multi-head self-attention (SW-MSA) \cite{liu2021swin-11} is applied to implement cross-window interactions to learn long-range dependency information. In this second ST block, every module is the same as the previous block except that input features are shifted by $\left \lfloor \frac{M}{2}\right \rfloor\times \left \lfloor\frac{M}{2} \right \rfloor$ before window partitioning. Using this alternating regular and shifting window partitioning way, it not only makes the ST block requires less computationally cost but also enables the ST block to have the cross-window interaction capability, thus capturing long-range dependencies in both spatial and temporal dimensions. Finally, the outputs of such ST blocks are downsampled by a convolutional layer with a stride of two, serving as the input of the next encoder stage.
 
\subsection{Decoder}
Symmetrically with the encoder part, the decoder of the USTN-DSC also consists of four stages with $D_n, n=0,1,2,3$, each of which in turn is followed by ST decoder block with a deconvolution layer for upsampling. The detailed structure of the ST decoder block is shown in \cref{fig:main}(c). We restore the missing frames in the decoding phase mainly by means of the conversion of the features from the keyframes extracted from the encoder part. For the missing video frames, they contain both slow moving objects, whose differences with keyframes are minimal, and objects with large motion, which need to be synthesized by inference of spatio-temporal relationship of the keyframes. In order to cope with these two different motion patterns for better restoration of missing video frames, we introduce the dual skip connections in the decoder section. First, we insert a corresponding multi-head cross-attention (MCA) after the regular and shifted windows-based multi-head self-attention. The MCA receives the features from the output of the previous decoding layer as query, and the features from the corresponding level of the encoder as key and value.
By querying the features at different scales and distances in the encoder part of the corresponding level, MCA enables the decoder to assist in the generation of certain fast-motion object features of the missing frames. Second, we design a TU module consisting of a 3D deconvolution layer with a kernel size of $(T-3)\times 3\times 3$ and stride size of $1\times 1\times 1$ to upsample the features generated by the encoder to obtain the features of the intermediate missing frames. (Note that the TU module in the skip connection shares weights, except for the one in the bottleneck section.) Then, the features at the timestamps of the corresponding keyframes are filled with the original features from the encoder. Finally, the combined features are added to the output of the corresponding level of the decoder in the form of residual. This operation can compensate for the lack of original detail features query in the cross-attention connection and can further facilitate the decoder to better restore the detail information of background and slow objects in video 
sequence.


\subsection{Loss Function}

We mainly consider the loss function from both appearance and motion aspects. First, we use the charbonnier loss \cite{lai2018fas-37t}, which compensates for the shortcomings of the $L_1$ and $L_2$ losses, to compute the RGB differences between the corresponding output frame $I_t$ and the real frame $\hat{I}_t$ for appearance constraint:
\begin{equation}
L_{cb}(\hat{I}_t,I_t)=\sqrt{{\left \| \hat{I}_t-I_t \right \|}^2+{\epsilon}^2},
\label{eq:eq3}
\end{equation}
where $\epsilon$ is set to $10^{-3}$ in our experiments. For the motion constraint, we introduce a simple and effective AFD loss:
\begin{equation}
\begin{aligned}
L_{fd}(\left\{\hat{I}_t\right\}^{T}_{t=1}&, \left\{{I}_t\right\}^{T}_{t=1})= \\
&\sum_{t=1}^{T-1}\sqrt{{\left \| {\left \|\hat{I}_t-\hat{I}_{t+1}\right \|}^2- {\left \|{I}_t-{I}_{t+1}\right \|}^2 \right \|}^2 +{\epsilon}^2}.
\end{aligned}
\label{eq:eq4}
\end{equation}
AFD loss directly promotes motion consistency by constraining the difference between the pixel of adjacent frames of the restored video sequence and the real video sequence. Compared with the computationally expensive optical flow constraint method, AFD loss is not only simple to compute but also has a comparable temporal constraint effect.
Finally, the overall loss function is given as follows:
\begin{equation}
L_{all}=L_{cb}+L_{fd}.
\label{eq:eq5}
\end{equation}

\subsection{Anomaly Detection on Testing Data}

During the testing phase, we take $T$-frames as the processing unit, but we do not use the error between each restored frame $I_t$ and the real frame $\hat{I}_t$ as its corresponding anomaly indicator. Because we find experimentally that the keyframes and frames adjacent to the keyframes have very slight errors with the real frames, even for anomalous events. Therefore, we take the $PSNR$ corresponding to the frame with the largest mean square error between the $T$-frames and the real frames as the anomaly detection indicator for this video sequence, formulated as follows:
\begin{equation}
\begin{cases}
t^*= \mathop{max}\limits_{1\leq t \leq T}\frac{1}{K}\sum^K_{i=0}(\hat{I}_{t,i}-I_{t,i})^2\\
PSNR(\hat{I}_{t^*},I_{t^*})=10log_{10}\frac{[max_{\hat{I}_{t^*}}]^2}{\frac{1}{K}\sum^K_{i=0}(\hat{I}_{t^*,i}-I_{t^*,i})^2}
\end{cases},
\label{eq:eq6}
\end{equation}
where $K$ is the total number of image pixels and $max_{\hat{I}_{t^*}}$ is the maximum value of image pixels. We assign the same $PSNR$ value to each frame of a processing unit for anomaly metric calculation. To quantify the probability of anomalies occurring, we normalize each $PSNR$, following work \cite{mathieu2015deep-56}, to obtain anomaly scores in the range  $[0,1]$:
\begin{equation}
S(t)=1-\frac{PSNR(\hat{I}_t, I_t)-min_t PSNR(\hat{I}_t, I_t)}{max_t PSNR(\hat{I}_t,I_t)-min_t PSNR(\hat{I}_t,I_t)}.
\label{eq:eq7}
\end{equation}

\section{Experiments}

\subsection{Experimental Setup}
\noindent\textbf{Datasets.} We evaluate the performance of our method on three classic benchmark datasets widely used in the VAD community. (1) Ped2 \cite{mahadevan2010anomaly-38}. It contains 16 training videos and 12 testing videos in fixed scenarios. Abnormal events include riding a bicycle, skateboarding, and driving a vehicle on the sidewalk. (2) Avenue \cite{lu2013abnormal-39}. It consists of 16 training videos and 21 testing videos with 47 abnormal events including throwing a bag, moving toward or away from the camera, and running on the sidewalk. (3) ShanghaiTech \cite{luo2017revisit-40}. It contains 330 training videos and 107 testing videos with 130 abnormal events, such as affray, robbery, fighting, etc., distributed in 13 different scenes.

\noindent\textbf{Evaluation Metric.} Following the widely used evaluation metrics in the field of VAD, we use the frame-level area under the curve (AUC) of receiver operation characteristic to evaluate the performance of our proposed method.

\noindent\textbf{Training Details.} In the training phase, we first resize each frame to the size of $256\times 256$, while the values of the pixels in all frames are normalized to $[0,1]$. Then, we use the Adam optimizer with $L_2$ and decoupled weight decay \cite{loshchilov2017decoupled-41} by setting ${\beta}_1=0.9$ and ${\beta}_2=0.99$ to train the USTN-DSC. The initial learning rate is set to $2\times10^{-4}$ and is gradually decayed following the scheme of cosine annealing. The length of the output video sequence $T$ is set to 9. The ST block depth $N$ of each stage in USTN-DSC is set to 6. Training epochs are set to 100, 150, 200 on Ped2, Avenue, and ShanghaiTech, respectively, with batch size set to 4. We train our model on a single NVIDIA RTX 3090 GPU.

\begin{table}
\centering
\arrayrulecolor{black}
\begin{tabular}{c|l|c|c|c} 
\hline
\multicolumn{2}{c|}{~~~~~~~~Methods}       & Ped2   & Avenue & SHTech  \\ 
\hline
\multirow{6}{*}{\rotatebox{90}{others}}       & SCL  \cite{lu2013abnormal-39}         & N/A & 80.9    & N/A       \\
                         & Unmasking  \cite{tudor2017unmasking-52}     & 82.2 & 80.6 & N/A       \\
                        
                         & AnomalyNet \cite{zhou2019anomalynet-54}  & 94.9 & 86.1 & N/A       \\
                         & DeepOC  \cite{wu2019deep-55}            & 96.9 & 86.6 & N/A       \\ 
                         & MPED-RNN \cite{morais2019learning-57}  & N/A & N/A & 73.4       \\ 
                         & Scene-Aware  \cite{sun2020scene-43}  & N/A & \underline{89.6} & \underline{74.7}       \\ 
                         
\hline
\multirow{8}{*}{\rotatebox{90}{R.}}      & Conv-AE \cite{hasan2016learning-8}  & 90.0 & 70.2 & 60.9    \\
                         & ConvLSTM-AE \cite{luo2017remembering-25}    & 88.1 & 77.0 & N/A       \\
                         & Stacked RNN \cite{luo2017revisit-40}    & 92.2 & 81.7 & 68.0    \\
                         & AMC  \cite{nguyen2019anomaly-53}           & 96.2 & 86.9 & N/A       \\
                           & MemAE \cite{gong2019memorizing-6}          & 94.1 & 83.3 & 71.2    \\
                         & CDDA \cite{chang2020clustering-44}           & 96.5 & 86.0 & 73.3    \\
                         & MNAD \cite{park2020learning-45}           & 90.2 & 82.8 & 69.8    \\
                         & Zhong et al. \cite{zhong2022cascade-46}           & \underline{97.7} & 88.9 & 70.7    \\ 
                         
\hline
\multirow{7}{*}{\rotatebox{90}{P.}} & FFP \cite{liu2018future-9}             & 95.4 & 84.9 & 72.8    \\
                         & AnoPCN \cite{ye2019anopcn-27}         & 96.8 & 86.2 & 73.6    \\
                         & MNAD \cite{park2020learning-45}          & 97.0 & 88.5 & 70.5    \\
                         & ROADMAP \cite{wang2021robust-48}        & 96.3 & 88.3 & \textbf{76.6}    \\
                         & MPN \cite{lv2021learning-47}            & 96.9 & 89.5 & 73.8    \\
                         & AMMC-Net \cite{cai2021appearance-49}            & 96.9 & 86.6 & 73.7    \\
                         & DLAN-AC \cite{yang2022dynamic-28}       & 97.6 & \textbf{89.9} & \underline{74.7}    \\
\hline
                         & \textbf{USTN-DSC}       & \textbf{98.1} & \textbf{89.9} & 73.8    \\
\hline
\end{tabular}
\arrayrulecolor{black}
 \caption{Quantitative comparison with the state of the art for anomaly detection. We measure the average AUC (\%) on Ped2 \cite{mahadevan2010anomaly-38}, Avenue \cite{lu2013abnormal-39}, and ShanghaiTech \cite{luo2017revisit-40} datasets. The comparison methods are listed in chronological order. ('R.' and 'P.' indicate the reconstruction and prediction tasks, respectively.)}
  \label{tab:tab1}
\end{table}

\begin{figure*}[t]
  \centering
   \includegraphics[width=0.7\linewidth]{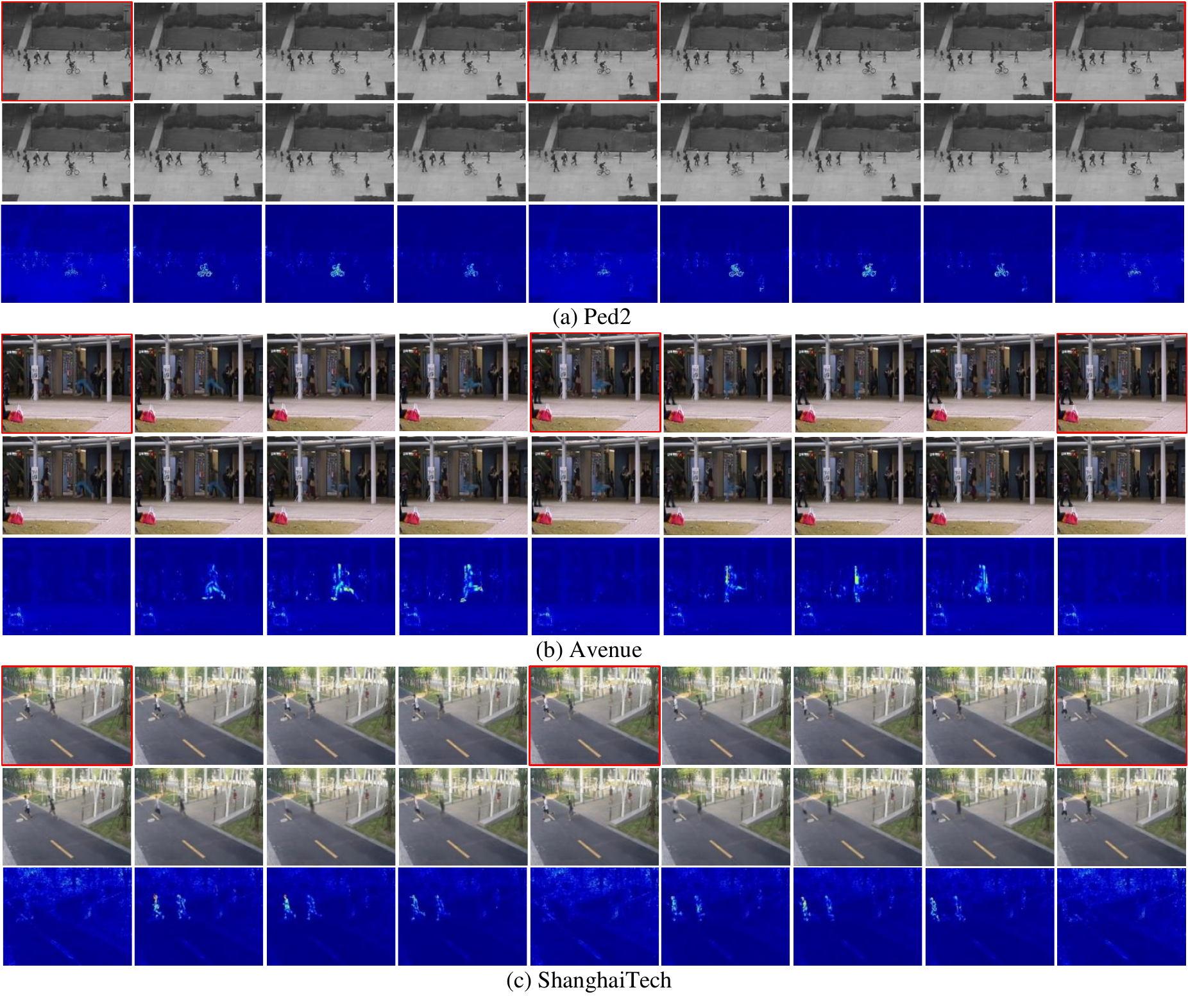}
   \caption{Examples of video events restoration on three datasets. For each dataset, the first row is the real video event sequence, the second row is the restoration results of our method based on keyframes, and the third row is the restoration error map. The frames with red borders are keyframes.}
   \label{fig:main_Result}
\end{figure*}

\subsection{Experiment Results}

\noindent\textbf{Comparison with Existing Methods.} We compare the performance of USTN-DSC with various state-of-the-art methods under different paradigms in \cref{tab:tab1}. It can be seen from \cref{tab:tab1} that the performance of our method on Ped2 and Avenue datasets achieve state-of-the-art compared to other methods and has a substantial improvement over the pioneer methods based on the deep learning reconstruction \cite{hasan2016learning-8} and prediction \cite{liu2018future-9} paradigms. This demonstrates that our method is a more effective modeling paradigm for learning normal behavior patterns to distinguish anomalies. For the ShanghaiTech dataset, the performance of our method does not achieve the optimum, but it is quite competitive compared with other methods. Because the ShanghaiTech dataset contains 13 different scenes, where the backgrounds and motion objects involved are quite complex and variable. This poses a higher demand on the ability of the model to learn the spatio-temporal relationships. However, we analyze the effect of different ST block depth $N$ on the model performance in sec.4.3 and demonstrate that USTN-DSC is able to further improve the model performance as the depth $N$ continues to increase.  In addition, USTN-DSC follows a simple U-Net architecture design. We do not employ other assists such as optical flow \cite{cai2021appearance-49}, adversarial training \cite{liu2018future-9}, extraction of the foreground objects \cite{morais2019learning-57}, or memory enhancement \cite{gong2019memorizing-6,park2020learning-45}. Nevertheless, compared to these well-equipped methods, USTN-DSC still surpassed them, which further validates the effectiveness of our method. 
\begin{figure}[t]
  \centering
   \includegraphics[width=1\linewidth]{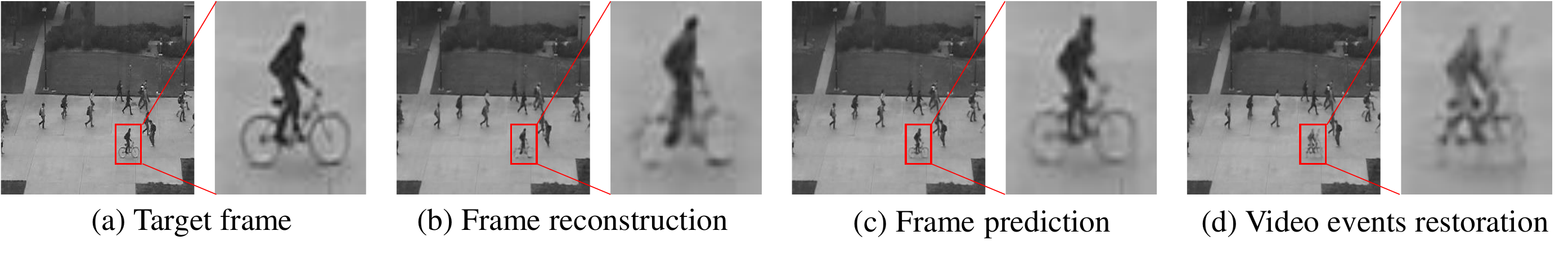}
   \caption{Compare the output results of USTN-DSC with reconstruction and prediction-based methods for anomalous samples.}
   \label{fig:R_P_Compare}
\end{figure}

\noindent\textbf{Qualitative Results.} We present the results of our method to restore video sequences based on keyframes of abnormal video samples on the ped2, avenue, and ShanghaiTech datasets, respectively, in \cref{fig:main_Result}. It can be observed that for the normal regions in the video frames, our method is able to restore them well, while drastic errors occur for abnormal event regions. Then, \cref{fig:R_P_Compare} shows the results of video frames restored by our method compared with the output of existing reconstruction-based \cite{park2020learning-45} and prediction-based methods \cite{liu2018future-9}. It can be seen that the frame (located in the middle of two keyframes) of the anomalous samples restored by our method have much large distortion and deformation errors in the anomalous region compared to the output of the previous two methods.
\begin{figure*}[t]
  \centering
   \includegraphics[width=0.7\linewidth]{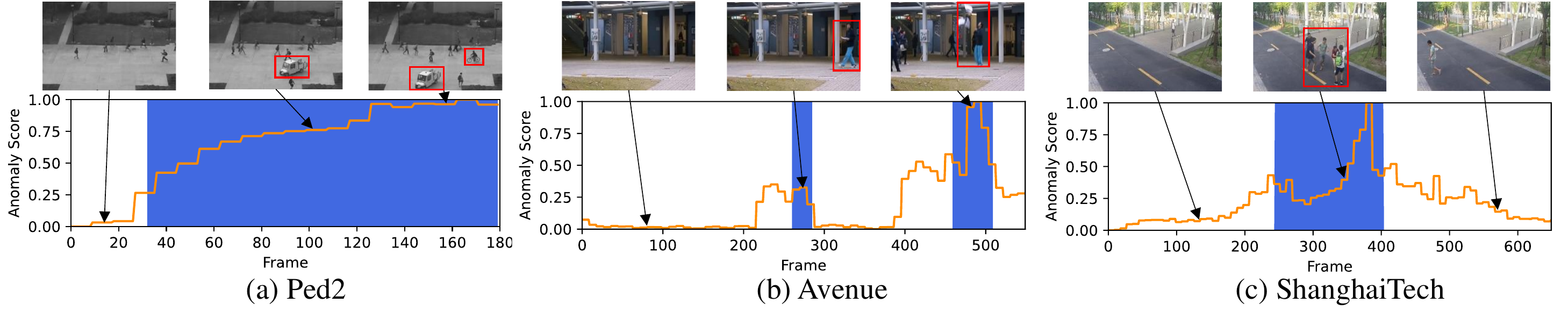}
   \caption{Anomaly score curves of several test samples of our method on three benchmark datasets.}
   \label{fig:AS}
\end{figure*}
This can further demonstrate that our approach, which considers the evolutionary relationship between appearance and motion over the long term, is able to effectively learn the more discriminative behavioral patterns in normal videos and thus be able to more accurately distinguish abnormalities. \cref{fig:AS} shows the anomaly score curves of some video clips on the three datasets. Obviously, there is a sharp jump in the anomaly score with the occurrence of anomalous events, and the anomaly curve returns to flat when the anomalous events disappear. This further demonstrates that our method has excellent sensitivity to anomalies and can effectively detect anomalous events.
 
\subsection{Ablation Study and Analysis}
In this section, we analyze the impact of several key components, network parameters, and loss functions on the performance of our method. Due to limited space, the analysis of the selection of different video sequence lengths $T$ is given in the supplementary materials. 

\noindent\textbf{Dual Skip Connections Analysis.} In \cref{tab:tab2}, we show the variation of the performance of our proposed network on the three datasets from without any skip connection to equipping two skip connections one by one. As we can see from \cref{tab:tab2}, for the encoder-decoder network without any skip connection, its performance is quite unsatisfactory. When adding these two skip connections in turn, the performance of our method is significantly improved. Especially for the addition of cross-attention skip connection, it boosts the performance on Ped2, Avenue, and ShanghaiTech datasets by 5.3\%, 4.7\%, and 3.8\%, respectively. This shows that the construction of dual skip connections plays a critical role in contributing to video event restoration. More detailed analysis can be referenced in the supplementary materials.

\begin{table}
\centering
\arrayrulecolor{black}
\begin{tabular}{c!{\color{black}\vrule}c!{\color{black}\vrule}c!{\color{black}\vrule}c} 
\hline
\multicolumn{1}{l!{\color{black}\vrule}}{~} & Ped2   & Avenue & ShanghaiTech  \\ 
\hline
w/o DSC                                          & 91.1 & 83.2 & 67.4        \\ 
\hline
w CAC                                     & 96.4(+5.3) & 87.9(+4.7) & 71.2(+3.8)        \\
\hline
w TUC                                     & 95.2(+4.1) & 85.4(+2.2) & 70.6(+3.2)        \\
\hline
w DSC                                     & 98.1(+7.0) & 89.9(+6.7) & 73.8(+6.4)        \\
\hline
\end{tabular}
\arrayrulecolor{black}
\caption{The AUC(\%) obtained by USTN-DSC with different skip connection configurations on Ped2 \cite{mahadevan2010anomaly-38}, Avenue \cite{lu2013abnormal-39} and ShanghaiTech \cite{luo2017revisit-40} datasets. (DSC: dual skip connections, CAC: cross attention connection, TUC: temporal upsampling connection)}
  \label{tab:tab2}
\end{table}

\begin{table}
\centering
\arrayrulecolor{black}
\begin{tabular}{c!{\color{black}\vrule}c!{\color{black}\vrule}c!{\color{black}\vrule}c} 
\hline
\multicolumn{1}{l!{\color{black}\vrule}}{~} & Ped2   & Avenue & ShanghaiTech  \\ 
\hline
w/o AFDL                                          & 97.2 & 88.5 & 71.5        \\ 
\hline
w AFDL                                     & 98.1(+0.9) & 89.9(+1.4) & 73.8(+2.3)        \\
\hline
\end{tabular}
\arrayrulecolor{black}
\caption{The AUC(\%) obtained by USTN-DSC with or without adjacent frames difference loss (AFDL) on Ped2 \cite{mahadevan2010anomaly-38}, Avenue \cite{lu2013abnormal-39} and ShanghaiTech \cite{luo2017revisit-40} datasets.}
  \label{tab:tab3}
\end{table}

\noindent\textbf{AFD loss Analysis.} \cref{tab:tab3} shows the performance differences on ped2, avenue, and ShanghaiTech datasets with and without AFD loss. It can be seen that the AFD loss has a performance boost on all three datasets, especially for the ShanghaiTech dataset, where it improves the AUC by 2.3\%. This is because the ShanghaiTech dataset involves 13 different scenarios with complex motion patterns of foreground objects that have a higher dependence on motion constraints. This demonstrates the effectiveness of our proposed AFD loss for motion constraints.

\noindent\textbf{ST Block Depth N Analysis.} As shown in \cref{tab:tab4}, we show the performance variation on the Ped2, Avenue, and ShanghaiTech datasets by setting $N$ to 2, 4, 6. From the \cref{tab:tab4}, we can find that the performance of USTN-DSC on all three datasets gradually improves as $N$ increases. Interestingly, for the Ped2 dataset, the performance improvement from increasing $N$ is quite slight, while the improvement is very obvious for the Avenue and ShanghaiTech datasets. This can be explained by the fact that the scenes in the Ped2 dataset are fixed and the motion patterns are relatively simple, so a shallow network can meet the modeling requirements. For the Avenue and ShangahiTech datasets, their scenes are more complex and diverse, and place higher demands on the modeling capabilities of the network. Due to hardware constraints, we are not attempting to set a larger $N$ currently. However, it can be expected that the performance on Avenue and ShanghaiTech datasets can be further improved as $N$ continues to increase.

\begin{table}
\centering
\arrayrulecolor{black}
\begin{tabular}{c!{\color{black}\vrule}c!{\color{black}\vrule}c!{\color{black}\vrule}c} 
\hline
\multicolumn{1}{l!{\color{black}\vrule}}{~} & Ped2   & Avenue & ShanghaiTech  \\ 
\hline
$N$=2                                          & 97.1 & 87.8 & 71.5        \\ 
\hline
$N$=4                                     & 97.7(+0.6) & 89.2(+1.4) & 72.6(+1.1)        \\
\hline
$N$=6                                     & 98.1(+1.0) & 89.9(+2.1) & 73.8(+2.3)        \\
\hline
\end{tabular}
\arrayrulecolor{black}
\caption{The AUC(\%) obtained by USTN-DSC with different ST block depth $N$ on Ped2 \cite{mahadevan2010anomaly-38}, Avenue \cite{lu2013abnormal-39}, and ShanghaTech \cite{luo2017revisit-40}.}
  \label{tab:tab4}
\end{table}

\section{Conclusions}
In this paper, we introduced a brand-new video anomaly detection paradigm that is to restore a video event based on keyframes. To this end, we proposed a novel model called USTN-DSC for video events restoration, where a cross-attention and a temporal upsampling residual skip connection are introduced to further assist in restoring complex dynamic and static motion object features in the video. In addition, we introduced a temporal loss function based on the pixel difference of adjacent frames to constrain the motion consistency of the video sequence. Extensive experiments on three benchmark datasets show that our method outperforms most existing state-of-the-art methods, demonstrating the effectiveness of our method.

{\small
\bibliographystyle{ieee_fullname}
\bibliography{PaperForReview}
}
\end{document}


\title{Video Event Restoration Based on Keyframes for Video Anomaly Detection\\ \emph{\textbf{(Supplementary Materials)}}}

\author{Zhiwei Yang$^1$, Jing Liu$^1$$^\dagger$\thanks{$^\dagger$Corresponding authors.}, Zhaoyang Wu$^1$, Peng Wu$^2$$^\dagger$, Xiaotao Liu$^1$\\ 
$^1$Guangzhou Institute of Technology, Xidian University, Guangzhou, China\\ 
$^2$School of Computer Science, Northwestern Polytechnical University, Xi'an, China\\
{\tt\small \{zwyang97, neouma, 15191737495\}@163.com, xdwupeng@gmail.com, xtliu@xidian.edu.cn}
}

\maketitle


\section{Analysis of the video sequence length $T$}
We show in \cref{tab:tab1} the variation of AUC on the Ped2 dataset for different values of $T$. We can infer that as $T$ decrease, the difference between video frames decreases, which reduces the difficulty of network modeling and increases the probability of an anomalous event being restored. The increase in the false negative rate reduces the overall performance. Conversely, as $T$ increases, the difference between video frames increases, which increases the difficulty of network modeling and decreases the probability of normal events being restored. The increase in the false positive rate also reduces the overall performance. When T is set to the middle value of 9, the best performance is obtained by USTN-DSC.

\begin{table}[htbp]
\centering
\begin{tabular}{c|c|c|c|c|c} 
\hline
$T$ & 5 & 7 & \textbf{9} & 11 & 13  \\ 
\hline
AUC & 96.4 & 97.2 & \textbf{98.1} & 94.5 & 94.3  \\
\hline
\end{tabular}
\caption{The AUC(\%) obtained by USTN-DSC for different values of $T$ on the Ped2 dataset.}
  \label{tab:tab1}
\end{table}

\section{Feature Extraction Module}
\label{sec:intro}
\cref{tab:tab2} shows the detailed architecture of the feature extraction module. Feature extraction module serves two main purposes. First, it takes advantage of the excellent local modeling ability of the convolutional neural network to capture the underlying local features, such as color, texture, edge, etc., which is beneficial to the restoration of detailed information of video frames in the decoding stage. Second, the feature extraction module can reduce the spatial resolution, thus effectively decreasing the computational effort of the network and accelerating the inference speed.
\section{Output Head}
The detailed architecture of the output head is shown in \cref{tab:tab3}. The main role of the output head is to upsample the low resolution features map output from the decoder to the target resolution. To better enhance the quality of the restored video frames, following work \cite{shi2016real-62}, we use the PixelShuffle operation for upsampling. 

\section{More Analysis of DSC}
\begin{figure*}[t]
  \centering
   \includegraphics[width=0.8\linewidth]{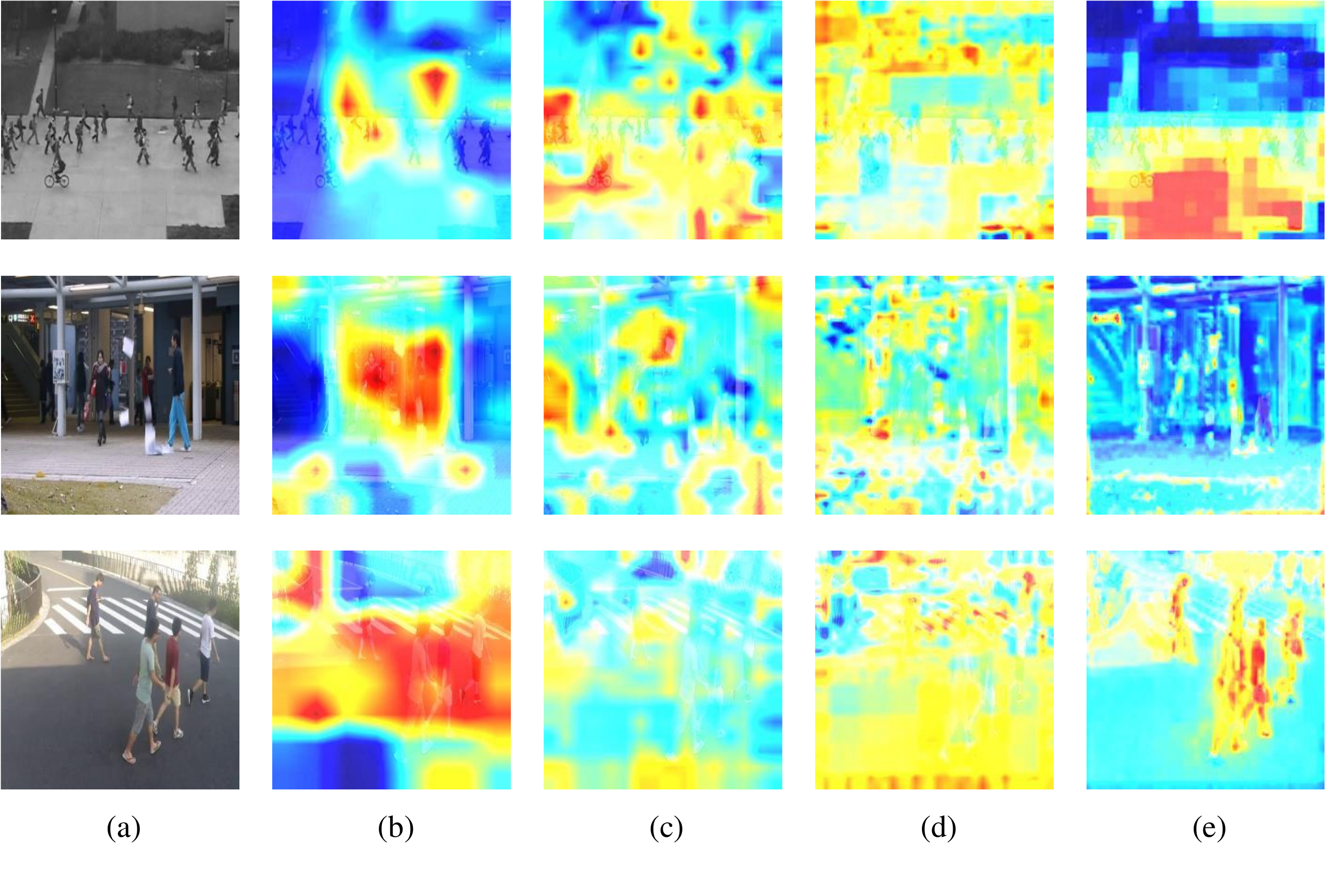}
   \caption{Visualization of attention maps of the across attention connection on some samples of  the Ped2, Avenue, and ShanghaiTech datasets. Column (a) denotes the ground-truth frame, and (b) $\sim$ (e) denote the attention maps generated by cross attention connections corresponding to the decoder $D_3 \sim D_0$ stages, respectively.}
   \label{fig:DSC-1}
\end{figure*}

\begin{figure*}[t]
  \centering
   \includegraphics[width=0.65\linewidth]{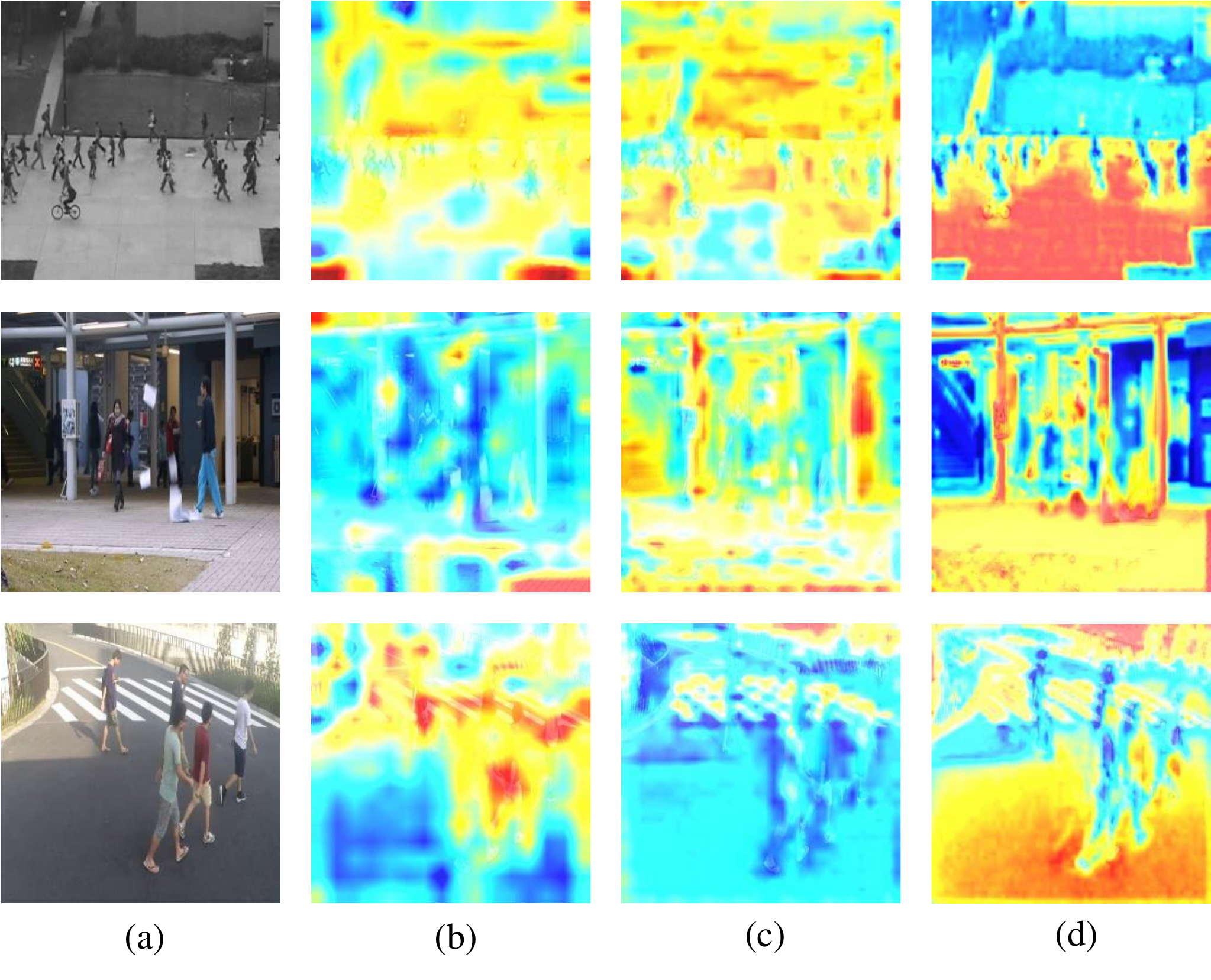}
   \caption{Visualization of the feature maps of temporal upsampling residual connection on some samples of the Ped2, Avenue, and ShanghaiTech datasets. Column (a) denotes the ground-truth frame, and (b) $\sim$ (d) denote the feature maps generated by temporal upsampling residual connections corresponding to the decoder $D_2 \sim D_0$ stages, respectively.}
   \label{fig:DSC-2}
\end{figure*}

Due to page limitations, we only quantitatively analyze the impact of DSC on model performance in the main paper. To demonstrate more intuitively the role of these two skip connections on the video event restoration task, we perform a qualitative analysis here. First, \cref{fig:DSC-1} visualizes the attention maps of cross attention connection on some samples of the Ped2, Avenue, and ShanghaiTech datasets. It can be observed from the \cref{fig:DSC-1} that the cross attention connection is mainly responsible for the transfer and transformation of dynamic objects features in the foreground. Further, we visualize the feature maps of the temporal upsampling residual connection in \cref{fig:DSC-2}, and it is obvious that this skip connection mainly serves the feature transfer and transformation of the background static objects. In addition, we can find that the cross attention connection and the temporal upsampling residual connection in different decoding stages are responsible for different foreground and background parts, which complement each other well. As shown in the quantitative analysis in Tab.2 of the main paper, the performance of the model not equipped with the DSC performs very poorly. The qualitative analysis here illustrates more intuitively that the design of the DSC plays a crucial role in the recovery of static and dynamic objects in video events, facilitating the USTN-DSC to more accurately model normal behavior patterns to better distinguish anomalies.


\section{Inference Speed}
In the inference stage, our method is implemented on a single NVIDIA RTX 3090 GPU on a machine with CPU core of i7-10700K@3.80Ghz and 32G memory. Our method takes on average  $5.3 \times 10^{-3}$ seconds (188FPS) to process each image of size $256 \times 256$.

\begin{table*}
\centering
\begin{tabular}{c|l|l} 
\hline
Layer name           & \multicolumn{1}{c|}{Structure}             & \multicolumn{1}{c}{Output size}      \\ 
\hline
Layer 1              & Conv2d ($3\times 3$) + BN + LeakyReLU             & (3, 256, 256, 64)  \\ 
\hline
Layer 2              & Conv2d ($3\times 3$) + BN + LeakyReLU             & (3, 256, 256, 64)  \\ 
\hline
Layer 3              & MaxPool2d ($2\times 2$)            & (3, 128, 128, 64)  \\ 
\hline
Layer 4              & Conv2d ($3\times 3$) + BN + LeakyReLU             & (3, 128, 128, 128)  \\ 
\hline
Layer 5              & Conv2d ($3\times 3$) + BN + LeakyReLU             & (3, 128, 128, 128)  \\ 
\hline
Layer 6              & MaxPool2d ($2\times 2$)            & (3, 64, 64, 128)  \\ 
\hline
Layer 7              & Conv2d ($3\times 3$) + LeakyReLU             & (3, 64, 64, 96)  \\ 
\hline
\end{tabular}
\caption{Network architecture of the feature extraction module.}
  \label{tab:tab2}
\end{table*}

\begin{table*}
\centering
\begin{tabular}{c|l|l} 
\hline
Layer name           & \multicolumn{1}{c|}{Structure}             & \multicolumn{1}{c}{Output size}      \\ 
\hline
Layer 1              & Conv2d ($3\times 3$)              & (9, 64, 64, 384)  \\ 
\hline
Layer 2              & PixelShuffle (2) + LeakyReLU             & (9, 128, 128, 96)  \\ 
\hline
Layer 3              & Conv2d ($3\times 3$)            & (9, 128, 128, 256)  \\ 
\hline
Layer 4              &  PixelShuffle (2) + LeakyReLU             & (9, 256, 256, 64)  \\ 
\hline
Layer 5              & Conv2d ($3\times 3$) + LeakyReLU             & (9, 256, 256, 64)  \\ 
\hline
Layer 6              & Conv2d ($3\times 3$)              & (9, 256, 256, 3)  \\ 
\hline
\end{tabular}
\caption{Network architecture of the output head.}
  \label{tab:tab3}
\end{table*}

{\small
\bibliographystyle{ieee_fullname}
\bibliography{Supplement_materials}
}